\documentclass[10pt,twocolumn,letterpaper]{article}

\usepackage{titling}
\usepackage{iccv}
\usepackage{times}
\usepackage{epsfig}
\usepackage{graphicx}
\usepackage{amsmath}
\usepackage{amssymb}
\usepackage{graphicx}
\usepackage{booktabs}
\usepackage{color}
\usepackage{xcolor}

\usepackage{multirow}
\usepackage{algorithm}
\usepackage{algorithmic}
\usepackage{eqparbox}
\usepackage{caption}

\usepackage{pifont}
\usepackage{dsfont}
\usepackage{gensymb}

\usepackage[accsupp]{axessibility}

\newcommand{\cmark}{\ding{51}}%

\usepackage[pagebackref=true,breaklinks=true,letterpaper=true,colorlinks,bookmarks=false]{hyperref}

\iccvfinalcopy

\newcommand{\ours}{ISCO}
\newcommand{\oursNI}{SCO}
\newcommand{\oursFull}{Iterative Superquadric reComposition of Objects}
\newcommand{\oursFullAcro}{\textbf{I}terative \textbf{S}uperquadric re\textbf{C}omposition of \textbf{O}bject views}

\newcommand{\myparagraph}[1]{\vspace{1pt}\noindent{\bf{#1}}}

\usepackage[capitalize]{cleveref}
\crefname{section}{Sec.}{Secs.}
\Crefname{section}{Section}{Sections}
\Crefname{table}{Table}{Tables}

\begin{document}

\title{Iterative Superquadric Recomposition of 3D Objects from Multiple Views}

\author{Stephan Alaniz\\
University of T\"ubingen 
\and
Massimiliano Mancini\\
University of Trento
\and
Zeynep Akata\\
University of T\"ubingen, MPI-IS
}

\maketitle

\begin{abstract}
Humans are good at recomposing novel objects, i.e. they can identify commonalities between unknown objects from general structure to finer detail, an ability difficult to replicate by machines.
We propose a framework, \ours, to recompose an object using 3D superquadrics as semantic parts directly from 2D views 
without training a model that uses 3D supervision. 
To achieve this, we optimize the superquadric parameters that compose a specific instance of the object, comparing its rendered 3D view and 2D image silhouette. 
Our \ours\ framework iteratively adds new superquadrics wherever the reconstruction error is high, abstracting first coarse regions and then finer details of the target object. With this simple coarse-to-fine inductive bias, \ours\ provides consistent superquadrics for related object parts, despite not having any semantic supervision. Since \ours\ does not train any neural network, it is also inherently robust to out-of-distribution objects.
Experiments show that, compared to recent single instance superquadrics reconstruction approaches, 
\ours\ provides consistently more accurate 3D reconstructions, 
even from images in the wild. Code available at {\small{\url{https://github.com/ExplainableML/ISCO}}}.

\end{abstract}
\vspace{-1em}

\section{Introduction}
\label{sec:intro}

\begin{figure}
     \centering
     \includegraphics[width=1.\linewidth]{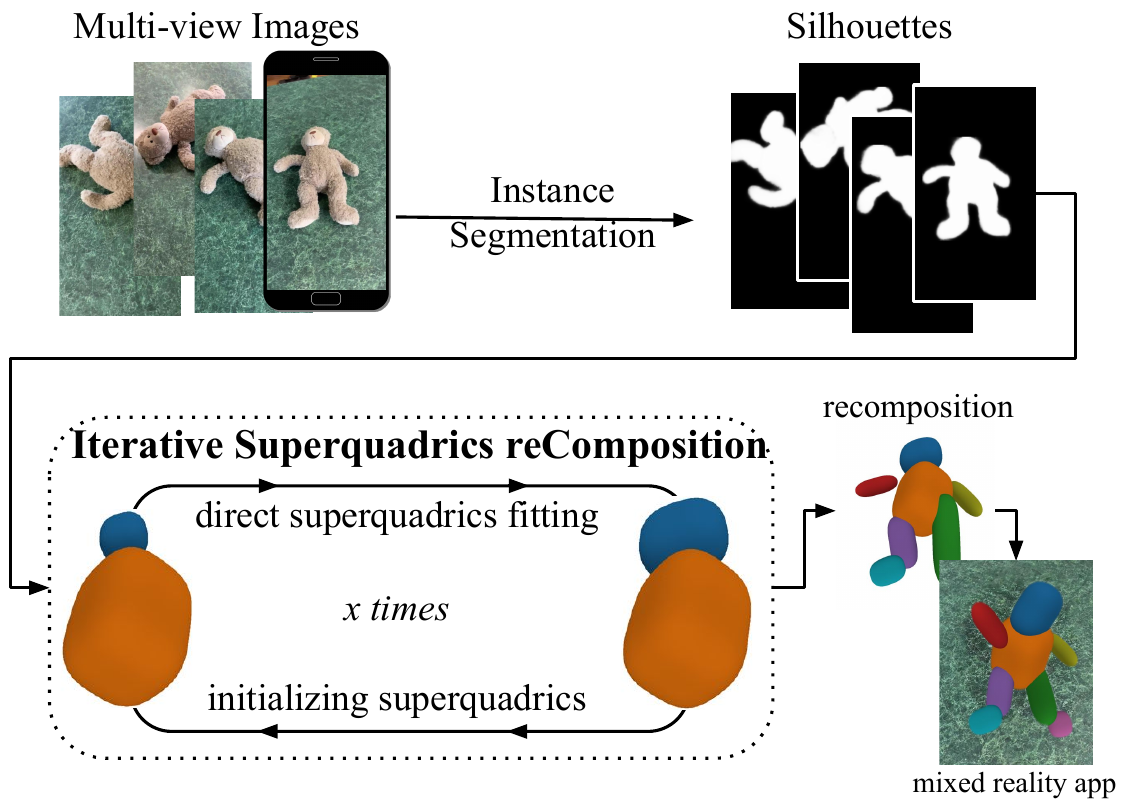}
     \caption{\ours\ recomposes 3D objects from real-world images using superquadrics. It acts on 2D silhouettes and not 3D models learned via a neural network, 
 placing new superquadrics iteratively into the scene. The final 3D abstraction and simple parts can then be used for mixed reality.}
     \label{fig:teaser}
 \end{figure}
 
Although the Jeff Koons sculpture ``Balloon Dog'' does not have a nose or teeth of a dog, we are able to recognize the dog when we look at the sculpture as a whole. This is because we can decompose the sculpture into semantically meaningful parts and 
recompose them 
in our minds to give it the name of a familiar object. 
Equipping machines with the same perceptual grouping capabilities would improve 
how they process and interact with the environment~\cite{pentland1987perceptual}.
 
In this context, decomposing 3D objects by means of 
simple 3D shapes 
\cite{pentland1986parts,chevalier2003segmentation} abstract away complex details of the object, providing a parsimonious representation 
useful in many 
applications, such as 
robotics~\cite{vezzani2017grasping} and virtual reality~\cite{moustakas2005geometry}.
Recent works addressed this task by training a 
a neural network on a collection of 3D shapes, 
 showing that semantically consistent part decompositions emerge, \ie networks predict the same semantic parts (\eg \textit{back}, \textit{legs}) across shapes (\eg \textit{chairs}, \textit{sofa})~\cite{tulsiani2017learning,paschalidou2019superquadrics,paschalidou2021neural}. 
 However,
 the data collection is costly in the real world, where the need for simple compositional representations is more evident. {At the same time, instance-based approaches \cite{liu2022robust,wu2022primitive} sidestep such need, but they require 3D inputs (\eg point clouds) and do not extract semantically relatable parts.} 

In this work we take a different perspective, proposing  \oursFull\ (\ours), an instance-level self-supervised algorithm that, given multiple 2D views of an object, recomposes its 3D shape using superquadrics~\cite{pentland1986parts,paschalidou2019superquadrics}, \textit{without} access to 3D data.  
\ours\ compares 
 real and synthesized 2D views via a differentiable rendering pipeline~\cite{mildenhall2020nerf}, 
directly optimizing the primitive parameters. 
To extract more precise abstractions, we perform an iterative recomposition of the object 
i) initializing primitives in regions with high reconstruction error; ii) fitting them by 
prioritizing local accuracy over global coverage, representing finer details as the number of primitives increases. Results on ShapeNet~\cite{ChangFGHHLSSSSX15} and ShapeNet-Part~\cite{Yi16} show that \ours\ achieves 
accurately reconstructs 3D shapes, identifying semantic parts much better than recent instance-based approaches\cite{liu2022robust,wu2022primitive}. Acting at instance-level and on multiple views, \ours\ can readily work on arbitrary objects and real-world images, as we show on Common Objects in 3D~\cite{reizenstein21co3d}. Fig.~\ref{fig:teaser} shows a simplified version of our pipeline. 

To summarize, our contributions are: i) we tackle the problem of 3D shape abstraction from multiple views, 
ii) we propose 
\ours, a self-supervised algorithm that 
iteratively fits superquadrics to unexplained input regions, directly optimizing their parameters 
via a differentiable rendering pipeline; 
iii) we show that \ours\ precisely reconstructs input shapes, is easily transferable to real-world applications and can better relate shape parts across different instances than previous instance-based approaches. 


\section{Related work}
\label{sec:related}

\myparagraph{3D Shape Decomposition.}
Among different ways to represent 3D shapes, voxel grids~\cite{liu2019point}, point clouds~\cite{qi2017pointnet}, triangular meshes~\cite{shewchuk2002delaunay}, and signed distance functions~\cite{park2019deepsdf} stand out. While they provide precise shape representations, their complexity results in limited interpretability. To address this issue works decomposed shapes into parts by using \eg supervised objectives on part annotations~\cite{mo2019partnet,qi2017pointnet,xu2021learning,chen2022latent}, zero/few-shot learning~\cite{luo2019learning,zhao20193d,sharma2022pri,naeem20223d}, language~\cite{koo2022partglot,corona2022voxel}, or unsupervised objectives~\cite{tulsiani2017learning,paschalidou2021neural,deng2020cvxnet,chen2019bae,chen2020bsp,wang2021learning,yao2021discovering,niu2022rim}. 

In the unsupervised setting, most works decomposed 3D shapes using simple volumetric primitives since compact shape representations better aid shape analyses and parsing~\cite{tulsiani2017learning,paschalidou2021neural,ren2022extrudenet}. 
For instance,~\cite{tulsiani2017learning} predicts cuboids given an input 3D volume.~\cite{paschalidou2019superquadrics} maps visual inputs (\eg single-image, point clouds) to compositions of superquadrics, achieving a better trade-off between compactness and fidelity.  
Subsequent works explored more complex decompositions~\cite{paschalidou2020learning}), scenes~\cite{kluger2021cuboids}, or atomic primitives (\eg convexes~\cite{deng2020cvxnet}, Gaussians~\cite{genova2019learning,genova2020local}, planes~\cite{chen2020bsp}, multi-tapered superquadrics~\cite{wu2022primitive}, algebraic surfaces~\cite{yavartanoo20213dias}, deformed shapes~\cite{paschalidou2021neural}). 

As in~\cite{paschalidou2019superquadrics,paschalidou2020learning} we still represent objects as composition of superquadrics, fitting a target shape with an unsupervised reconstruction objective. More closely related to us are early works fitting superquadrics to single instances given range data~\cite{leonardis1997superquadrics,solina1990recovery}, and especially more recent ones using point clouds as input~\cite{wu2022primitive,liu2022robust}. However, we do not need 3D inputs~\cite{wu2022primitive,liu2022robust}, nor a  training set as in~\cite{paschalidou2019superquadrics}. 
Instead, we directly obtain abstractions from multiple views  iteratively and 
via a differentiable rendering pipeline. Despite not using explicit 3D input, we show that \ours\ better reconstructs the target shape than \cite{wu2022primitive,liu2022robust} and provides semantic parsing results much closer to those of specific methods~\cite{chen2019bae,niu2022rim}. 

\myparagraph{Neural Radiance Fields.}
Since the seminal paper of Mildenhall et al.~\cite{mildenhall2020nerf}, NeRF has been the de-facto standard to synthesize novel views of a target scene. Practically, NeRF learns an implicit neural scene representation by mapping 3D coordinates and viewpoints to corresponding color and density values. This mapping is instance-specific and exploits a differentiable rendering pipeline to enforce consistency between the rendered and target views. 

Subsequent works improved NeRF in multiple aspects, \eg number of views~\cite{yu2021pixelnerf,muller2022autorf,xu2022sinnerf}, relative constraints~\cite{pumarola2021d,martin2021nerf}, surface reconstruction~\cite{oechsle2021unisurf}synthesis quality~\cite{barron2021mip,guo2022nerfren}, scale~\cite{tancik2022block,turki2022mega}, and computational efficiency~\cite{reiser2021kilonerf,garbin2021fastnerf,hu2022efficientnerf}. In this regard, recent works~\cite{sun2022direct,liu2020neural,yu2021plenoctrees,xu2022point} tried to replace the implicit scene representation of NeRF with explicit counterparts, using signed distance function~\cite{wang2021neus}, voxel grids~\cite{sun2022direct}, octrees~\cite{liu2020neural,yu2021plenoctrees} or points~\cite{xu2022point}. While our method shares the same objective of fitting an explicit 3D representation to 2D views, we tackle the problem from a different perspective, \ie recompose a target object with simple volumetric shapes, representing its parts. 
We achieve this by revisiting the original NeRF formulation, iteratively fitting superquadrics to the available 2D views. 

Finally, our work is related to multi-view 3D reconstruction. This classical computer vision problem has various solutions, from Structure from Motion~\cite{schonberger2016structure,ozyecsil2017survey}, and SLAM~\cite{fuentes2015visual} methods, to more recent multi-view stereo~\cite{campbell2008using,schonberger2016pixelwise} and learning-based~\cite{wang2021multi,xie2019pix2vox,yariv2020multiview} ones. While \ours\ shares the same principles, our objective is different: providing an abstract composition of the target shape.

\begin{figure*}
    \centering
    \includegraphics[width=1.0\linewidth]{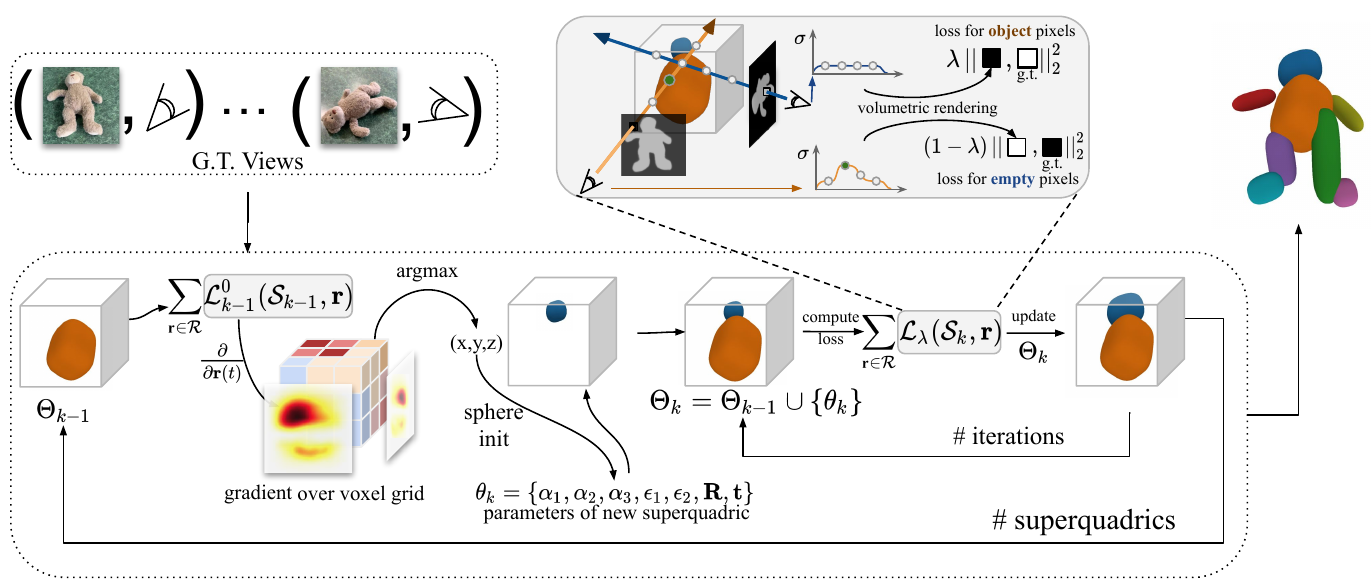}
    \caption{\textbf{\oursFull\ (\ours)} recomposes 3D objects from 2D views. We iteratively initialize a new superquadric to the region where the estimated error is the highest (left) and optimize its shape parameters using differential rendering (right). The output is a 3D reconstruction of the object with  superquadrics composing its parts. 
    } 
    \label{fig:method}
\end{figure*}

\section{Superquadric Recomposition from 2D views} 
After formalizing our setting in Sec.~\ref{sec:problem}, 
we describe our building blocks 
in Sec.~\ref{sec:primitives}. Finally, we describe our method 
that iteratively initializes and fits superquadrics 
in Sec.~\ref{sec:ours}. 

\subsection{Problem definition}
\label{sec:problem}
Our goal is to represent the volume of a static 3D object as a composition of simple volumetric shapes. When the 3D shape of an object is known, we can supervise this process by measuring how well the parts cover 
the ground-truth volume. 
While previous works~\cite{tulsiani2017learning,paschalidou2019superquadrics,paschalidou2020learning,paschalidou2021neural} assume the availability of such dense 3D data for supervision, in this work we relax this assumption and tackle the problem using images from multiple viewpoints around the object. 

Formally, we have as input a set of images $\mathcal{I}$ depicting the target object from different known viewpoints. Our objective is to obtain a set of 3D parts $\mathcal{S}$ that best reconstructs 
the shape of the object as a whole. Note that, similarly to previous works~\cite{tulsiani2017learning,paschalidou2019superquadrics,genova2019learning, deng2020cvxnet,liu2022robust,wu2022primitive}, our aim is to recompose the 3D shape of the target object, ignoring its particular texture or background. To achieve this,  
we supervise our model with {silhouettes} of the object extracted from the 2D views, \eg using {an out-of-the-box} segmentation algorithm. 

{Without a ground-truth 3D representation, we cannot directly compare the compositions of our parts, \ie superquadrics, to the target shape}. 
However, we can 
use a ray-marching algorithm similar to
~\cite{mildenhall2020nerf} to render 
silhouettes
of our superquadrics, in a fully differentiable manner, 
comparing the rendered silhouettes with the real ones 
in $\mathcal{I}$. Differently from~\cite{mildenhall2020nerf}, we do not train a neural network, but we directly update the superquadric parameters. In the following, we describe how we parameterize and render them. 

\subsection{Differentiable Superquadrics Rendering}
\label{sec:primitives}
\myparagraph{Superquadric representation.} The main building block of our algorithm is the simple volumetric 3D shape  
that we use to compose and represent the target object. We employ superquadrics~\cite{barr1981superquadrics} as primitives due to the variety of shapes they can represent with a simple parameterization that, 
as in~\cite{paschalidou2020learning,liu2022robust}, 
has the implicit surface function $f$:
\begin{equation}
    f(\mathbf{x}; \theta) = \left ( \left ( \frac{x_1}{\alpha_1} \right )^{\frac{2}{\epsilon_2}} + \left ( \frac{x_2}{\alpha_2} \right )^{\frac{2}{\epsilon_2}} \right ) ^{\frac{\epsilon_2}{\epsilon_1}} + \left ( \frac{x_3}{\alpha_3} \right )^{\frac{2}{\epsilon_1}}
\end{equation}
where $\theta$ are the parameters describing a superquadric, with $\{\alpha_1, \alpha_2, \alpha_3\}$ being the scale along the coordinate axes, and $\{\epsilon_1, \epsilon_2\}$ the shape  in its canonical form. A point $\mathbf{x}$ lies on the surface of the superquadric if  $f(\mathbf{x}; \theta) = 1$, inside if $f(\mathbf{x}; \theta) < 1$, and outside if 
$f(\mathbf{x}; \theta) > 1$. 

To make the superquadrics more expressive, we define additional rigid transformation parameters:
\begin{equation}
    \mathcal{T}(\mathbf{x}; \theta) = \mathbf{R}(\theta)\mathbf{x} + \mathbf{t}(\theta)
\end{equation}
where $\mathbf{R}(\theta)$ is a 3D rotation matrix in Euler formulation and $\mathbf{t}(\theta)$ is a translation vector. Given a superquadric $p$, its set of parameters is thus $\theta_p=\{\alpha_1, \alpha_2, \alpha_3, \epsilon_1, \epsilon_2, \mathbf{R}, \mathbf{t}\}$. For simplicity, we will denote as $\hat{f}(\mathbf{x};\theta)$ the superquadric after applying the global transformations. 

\myparagraph{Rendering superquadrics via ray-marching.}
To render superquadrics, 
we define a density function $\sigma: \mathbb{R}^3 \rightarrow [0, 1]$
\begin{equation} \label{eq:sq_density}
    \sigma(\mathbf{x}; \theta) = \mathtt{sigmoid} \left (\gamma (1 - \hat{f}(\mathbf{x}; \theta) ) \right)
\end{equation}
where 
$\gamma$ is a scalar determining the slope of the surface boundary. 
With a large $\gamma$, $\sigma(\mathbf{x}; \theta)$ behaves like a step function that is $1$ inside the superquadric and $0$ outside, 
accurately reproducing its density. However, when fitting the superquadrics, we seek a smooth transition of $\sigma(\mathbf{x}; \theta)$ to obtain well-behaved gradients, thus $\gamma$ is a hyperparameter.

With Eq.~\eqref{eq:sq_density}, we can now render superquadrics in 2D. 
Given a superquadric $s_k \in \mathcal{S}$, we adopt the volume rendering technique developed in~\cite{mildenhall2020nerf} to obtain the density value along a given camera ray $\textbf{r}(t) = \textbf{o} + t \textbf{d}$, with \textbf{o} being the origin and \textbf{d} its direction. 
We estimate the expected density of $s_k$ between near and far bounds $t_n$ and $t_f$ as
\begin{equation}
\label{eq:rendering}
    D(\textbf{r}, s_k) = \int_{t_n}^{t_f} T(t) \sigma(\textbf{r}(t), \theta_k)dt
\end{equation}
where $T(t) = \text{exp} \left (- \int_{t_n}^t \sigma(\textbf{r}(s), \theta_k) ds \right )$. 
Since we want to reconstruct 2D object silhouettes, we omit color values and directly use the density along the camera ray as gray-scale pixel intensity. To numerically estimate this continuous integral, we use the same stratified sampling approach as in~\cite{mildenhall2020nerf}.  Note that, differently from~\cite{mildenhall2020nerf}, our 
$\sigma$ is not computed via a neural network, but directly computed from Eq.~\eqref{eq:sq_density} using the parameters of the superquadrics. 

\subsection{Iterative Recomposition with Superquadrics}
\label{sec:ours}
With Eq.~\eqref{eq:rendering}, we can render our abstract representation given a viewpoint and use the discrepancy between rendered and real 2D views to fit the superquadrics to the target object. 
One straightforward approach is to randomly initialize a set of superquadrics and jointly optimize their parameters. 
However, this 
tends to find local optima, \eg not covering all parts of the shape and/or the splitting a single part with 
multiple superquadrics (see Sec.\ref{sec:exp}). 

To address this issue, we 
propose an iterative pipeline, adding one superquadric at the time. 
We name our algorithm \oursFullAcro\ (\ours). Each iteration in \ours\ has two steps: initialization of a new superquadric and its fitting. Fig~\ref{fig:iter_example} shows a simplified overview of the iterative fitting of superquadrics to a  teddy bear, visualizing both the initialization step and the fitting process, with the final object representation being composition of the superquadrics in $\mathcal{S}_K$. 
In the following, we describe our objective function to fit a superquadric to ground-truth views, and how we use the same to initialize a new superquadric at the beginning of each iteration. 

\myparagraph{Objective function.}
Given a specific viewpoint, the rendered view of our superquadrics $\mathcal{S}_k$ at step $k$ should match the ground-truth one in $\mathcal{I}$.  
Thus, our objective function is
\begin{equation}
    \label{eq:loss}
    \mathcal{L}_k = \sum_{\textbf{r} \in \mathcal{R}} ||D(\textbf{r}, \mathcal{S}_k) - I(\textbf{r})||_2^2.
\end{equation}
where \textbf{r} is a camera ray in the set of all rays $\mathcal{R}$, $I(\textbf{r})$ is the ground-truth (silhouette) value of the pixel for ray $\textbf{r}$, and $D(\textbf{r},\mathcal{S}_k) = \min(\sum_{i=1}^k D(\textbf{r},s_i), 1)$. 
While $\mathcal{L}_k$ gives the same importance to pixels inside and outside the object, we found that it is beneficial to weight differently the two cases. We thus define our final objective at step $k$ as
\begin{equation}
    \label{eq:l-loss}
    \mathcal{L}_k^\lambda = \sum_{\textbf{r} \in \mathcal{R}} \mathcal{L}_\lambda(\mathcal{S}_k, \mathbf{r})=\sum_{\textbf{r} \in \mathcal{R}} w_\lambda(\textbf{r})||D(\textbf{r}, \mathcal{S}_k) - I(\textbf{r})||_2^2.
\end{equation}
where $w_\lambda(\textbf{r})$ is a weighting function with value $\lambda$ if the ray is outside the object (\ie $I(\textbf{r})=0$) and $1-\lambda$ if it is inside (\ie $I(\textbf{r})>0$). 
With $\lambda$ we can achieve a better trade-off between covering large parts of the objects (\ie $0<\lambda<0.5$) and/or focusing on local fitting (\ie $\lambda>0.5$). 
In the experiments, we follow the latter strategy, setting $\lambda=0.6$, analyzing various $\lambda$ values in the supplementary. 

\myparagraph{Superquadrics fitting via direct optimization.}
The objective in Eq.~\eqref{eq:l-loss} measures the discrepancy between rendered superquadrics and real views of the target object, allowing us to directly optimize the shapes in $\mathcal{S}_k$. This is possible thanks to i) the simple parameterization of superquadrics and ii) the differentiable rendering of Eq.~\eqref{eq:rendering}. Given the set of superquadrics $\mathcal{S}_k$ with parameters $\Theta_k = \{\theta_1,\cdots,\theta_k\}$, we compute the gradient $\partial{\mathcal{L}_k^\lambda}/\partial{\Theta_k}$, and update the parameters via gradient descent:
\begin{equation}
    \label{eq:primitives-optimization}
    \Theta_k \leftarrow \Theta_k+\mathtt{optim}(\frac{\partial{\mathcal{L}_k^\lambda}}{\partial{\Theta_k}})
\end{equation}
where $\mathtt{optim}$ is an optimizer computing the specific update step from the gradient, 
\eg Adam~\cite{kingma2015adam} in our experiments. 

\myparagraph{Initialization by estimating missing parts.}
The loss in Eq.~\eqref{eq:l-loss} pushes the rendered and real views to match, but 
it does not guarantee that the superquadrics reconstruct all parts of the object after convergence. To encourage full coverage, we instantiate new superquadrics around object 
parts 
that have not been explained by any superquadric yet. 

To estimate the position of missing parts, we use a dense voxel grid $G \in \mathbb{R}^{N\times N\times N}$, with resolution $N$, and we propagate rendering errors to the {voxel grid}. 
Specifically, we compute the gradient of $\mathcal{L}_{k-1}^{\lambda=0}$ w.r.t. the superquadrics density at every voxel $V_{\mathbf{g}} = \sigma(\mathbf{g}; \theta)$ with $\mathbf{g}\in G$. To do so, we render the superquadrics using ray marching as in Eq.~\eqref{eq:rendering}, but instead of using the density value along the ray $\sigma(\textbf{r}(t), \theta_k)$ we trilinearly interpolate the density from the voxel grid $V$ at the ray points $\textbf{r}(t)$, the result of which we denote as $K_{\textbf{r}(t)}$.
We then propagate the gradients from $\mathcal{L}_{k-1}^{\lambda=0}$, where $\lambda=0$ to consider only rays hitting the target object, to each $V_{\mathbf{g}}$ as 
\begin{equation}
    \frac{\partial {\mathcal{L}_{k-1}^0}}{\partial V_{\mathbf{g}}} = \sum_{\textbf{r}\in \mathcal{R}} \sum_{\textbf{r}(t) \in \textbf{r}} \frac{\partial \mathcal{L}_{k-1}^0}{\partial{K_{\textbf{r}(t)}}} \prod_{i=1}^3 \max(0, 1 - \frac{|x_i^{(\mathbf{r}(t))} - x_i^{(\mathbf{g})}|}{l})
\end{equation}
where $[x_1^{(\mathbf{p})}, x_2^{(\mathbf{p})}, x_3^{(\mathbf{p})}]$ are the three-dimensional coordinates of point $\mathbf{p}$ and $l$ is the uniform distance between voxels in the grid. 
We provide a more detailed derivation in the supplementary.

Since the gradients over the voxel grid are an estimate of the regions with highest reconstruction errors, \ie the higher the gradient and the higher the discrepancy between rendered and ground-truth views, we can use them 
to perform an informed initialization of a new superquadric. 
To initialize a new superquadric $s_{k}$, we first smooth $G$ with a Gaussian kernel, and then assign the translation parameters in $\theta_{k}$  
to the coordinates of the maximum gradient value in the voxel grid, \ie $\textbf{t}(\theta)_{k} \leftarrow \text{arg} \max_{\mathbf{g}}  {\partial {\mathcal{L}_{k-1}^0}}/{\partial V_{\mathbf{g}}}$. We initialize the other parameters so that $s_{k}$ is a sphere, adding it to 
the set $\mathcal{S}_k$, \ie $\mathcal{S}_k=\mathcal{S}_{k-1} \cup \{s_{k}\}$. 
Optimizing the superquadric starting from a region of high error ensures that we cover missing object parts, improving the quality of the abstraction as the number of superquadrics increases. To instantiate the first superquadric, we use the rendering error over an empty scene, \ie 
$\mathcal{S}_0 = \emptyset$.

\section{Experiments}
\label{sec:exp}

In this section, we evaluate \ours\ both quantitatively and qualitatively. 
We first describe the experimental setup, and then compare our method with related instance-based superquadrics fitting algorithms using point clouds \cite{liu2022robust,wu2022primitive}. 
Furthermore, we ablate important aspects of our model, 
such as the number of views, superquadrics, and 
its iterative structure. Finally, we show results on semantic shape parsing and application on real world objects.

\myparagraph{Datasets.}  To compare \ours\ with previous works, we use ShapeNet~\cite{ChangFGHHLSSSSX15} with the splits of~\cite{ChoyXGCS16}, containing synthetic 3D models of 13 different classes, totaling more than 43k different instances. Having access to ground-truth shapes allows us to evaluate the accuracy of our reconstructions, even if we do not use 3D data in \ours. To obtain 2D views, we sample 4, 8 or 16 different camera positions randomly from all angles around the object from which we render the respective silhouettes images. 
{The camera poses are part of the input to \ours\ and are required for the differentiable rendering step.} 
We use the same procedure for evaluating semantic parsing performance, using the part annotations for ShapeNet provided in ShapeNet-part~\cite{Yi16} and considering the same objects, parts, and evaluation protocol in~\cite{niu2022rim}.

Finally, to test real-world applicability of \ours, we use the Common Objects in 3D (CO3D)~\cite{reizenstein21co3d} dataset. It contains multi-view images of around 19k objects from 50 MS-COCO~\cite{lin2014microsoft} categories annotated by their corresponding camera viewpoints. CO3D illustrates a realistic pipeline, where we start from natural images, perform instance segmentation with PointRend~\cite{KirillovWHG20} to obtain object masks, and apply \ours\ as 3D abstraction method. For each scene we uniformly sample 16 random views along the camera trajectory from the video around the object.

\myparagraph{Implementation Details.} 
We tune the hyperparameters of our instance-based optimization process on the training set of ShapeNet, using the same set of hyperparameters for both the ShapeNet test set as well as ShapeNet-part, and CO3D. During the optimization procedure, we use 128x128 images, sample 500 rays per image, and perform 250 update steps before adding a new superquadric. Unless otherwise stated, we set $\lambda = 0.6$ 
and a maximum number of 10 superquadrics. Due to the iterative nature of \ours, we obtain recompositions of the objects 
not only for 10 superquadrics, but any intermediate number as well with a single run. 
Our implementation is based on PyTorch3D~\cite{ravi2020pytorch3d}.

\myparagraph{Baselines.} 
Since we perform instance-based 3D reconstruction, we compare our model with two recent state-of-the-art approaches, EMS~\cite{liu2022robust} and NBP~\cite{wu2022primitive}.  EMS~\cite{liu2022robust} is a robust probabilistic algorithm for superquadrics recovery from point clouds, optimizing the superquadric parameters via the trust-region reflective algorithm \cite{coleman1996interior} and avoiding local optima by searching for alternative parameters sets. 
NBP~\cite{wu2022primitive} extends \cite{liu2022robust} by fitting a mixture of tapered superquadrics~\cite{barr1984global} to the input point-cloud, alternating between points clustering and superquadrics optimization. Note that both algorithms use a point cloud as input, while we rely on multiple views. Since there are various algorithms for generating point clouds from multiple views, we ensure a fair comparison by using a \textit{point cloud} sampled from the original mesh surface as input to EMS and NBP in the experiments on ShapeNet and ShapeNet-parts. Since in CO3D we do not have a ground-truth shape, we use \cite{liu2019soft} to fit a mesh from 16 views, sampling the relative point cloud. Additionally, we include NeuS~\cite{wang2021neus}, a neural implicit surface method that does not perform abstraction or object decomposition, but models the object implicitly with a neural network. NeuS serves as reference for non-abstract instance-based implicit shape representations. We also report results without the iterative procedure but initializing randomly all superquadrics (following the error of the empty scene $\mathcal{S}_0$) and optimizing them all at once (\oursNI).

\subsection{Shape Reconstruction on ShapeNet}

 {
\renewcommand{\arraystretch}{1}
 \begin{table}
\resizebox{1.0\columnwidth}{!}{
    \centering
    \begin{tabular}{l|cc|cc||c}
        Input &\multicolumn{2}{c|}{Point Cloud}&\multicolumn{3}{c}{2D Views}\\
        Method & EMS~\cite{liu2022robust} & NBP~\cite{wu2022primitive} & \oursNI & \ours & NeuS~\cite{wang2021neus}\\
        \midrule
        airplane & 0.148 & 0.589 & 0.626       & \bf{0.696} & 0.680\\
        bench    & 0.066 & 0.443 & 0.489       & \bf{0.499} & 0.630\\
        cabinet  & 0.598 & 0.620 & 0.599       & \bf{0.744} & 0.784\\
        car      & 0.546 & 0.543 & 0.610       & \bf{0.703} & 0.681\\ 
        chair    & 0.148 & \bf{0.595} & 0.526  & 0.549 & 0.706\\
        display  & 0.358 & \bf{0.635} & 0.629  & 0.631 & 0.738\\
        lamp     & 0.174 & 0.520 & 0.404       & \bf{0.531} & 0.593\\
        speaker  & 0.599 & 0.643    & 0.576    & \bf{0.756} & 0.776\\
        rifle    & 0.261 & 0.556    & 0.584    & \bf{0.692} & 0.736\\
        sofa     & 0.319 & 0.606     & 0.601   & \bf{0.694} & 0.780\\
        table    & 0.083 & 0.568 & 0.487       & \bf{0.576} & 0.610\\
        phone    & 0.665 & 0.753    & 0.729    & \bf{0.791} & 0.707\\
        vessel   & 0.322 & 0.578      & 0.652  & \bf{0.663} & 0.693\\
        \midrule                                                                                           
        mean     & 0.330 & 0.588  & 0.578      & \bf{0.656} & 0.701\\
    \end{tabular}
    }
    \caption{\textbf{Reconstruction on ShapeNet.} We report IoU of the object volume (higher is better). EMS and NBP use point cloud as input, ours and NeuS use 16 random views. 
    }
    \label{tab:quan_shapenet}
\end{table}
 }

In Tab.~\ref{tab:quan_shapenet}, we show the performance of \ours,  EMS, NBP, and \oursNI\ on 3D reconstruction on ShapeNet, where we use the same evaluation code of~\cite{paschalidou2020learning} to calculate the IoU of our superquadric reconstructions and the ground truth object meshes. \ours\ 
 consistently outperforms all others in average (\ie 0.656 vs 0.588 of NBP), despite using 2D views and not point clouds as input. \ours\ achieves the best results both in simpler objects (\eg \textit{speaker}, +0.11 on NBP) and more complex ones (\eg +0.085 on NBP on \textit{vessel}). Note also that NBP uses the more flexible tapered superquadrics but still achieves lower reconstruction accuracy than \ours.

 Interesting is also the gap with EMS and \oursNI\ that use the same superquadric parametrization and thus the same representation capacity. Our non-iterative pipeline (\oursNI), despite using the same input and optimization procedure, achieves less accurate reconstructions (\ie -0.08 mIoU on average across all categories), showing the benefit of our iterative procedure. The large gap with EMS (\ie 0.330 vs 0.656 on average) clearly shows how our initialization strategy and optimization procedure provide more precise feedback than existing probabilistic approaches. Overall, \ours\ gets close to NeuS on average (0.656 vs. 0.701) and even surpasses it in some categories, such as airplane, car, and phone. Due to the limited amount of views, the inductive biases of superquadrics in \ours\ can help modeling uncertain parts of the object which are not exposed by  any view.

\subsection{Analyzing and ablating \ours\ on ShapeNet}
\label{sec:ablation}
In this section, we dissect our \ours\ method to motivate its components and understand both its strengths and shortcomings. We analyze how the choice of the number of superquadrics and multi-view images affect its performance. Additionally, we study the impact of the iterative fitting process of \ours\ to highlight its importance in inferring accurate superquadrics recompositions. We analyze the hyperparameter $\lambda$ in the supplementary.

\begin{figure}
\centering
    \centering
    \includegraphics[width=0.95\linewidth]{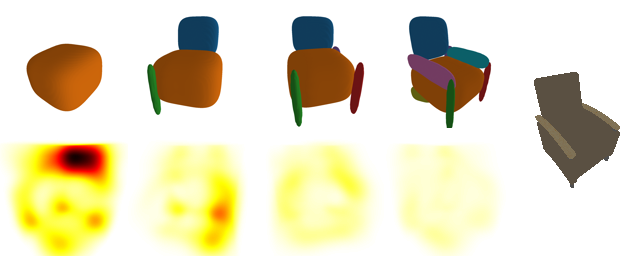}
    \includegraphics[width=0.95\linewidth]{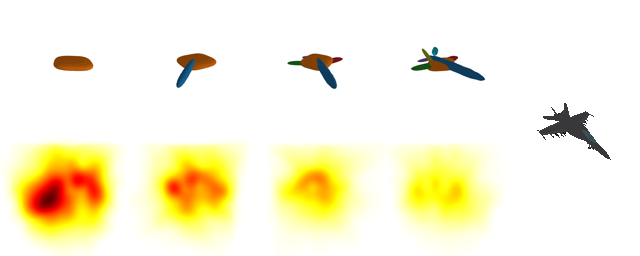}
    \includegraphics[width=0.95\linewidth]{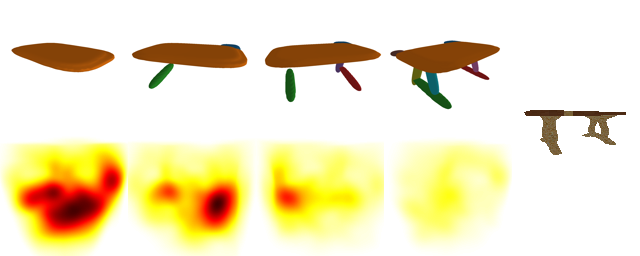}
    \captionof{figure}{\textbf{Visualizing \ours\ steps.} \ours\ recomposes the object with a superquadric at a time, going from coarse to fine (top). The gradients of the reconstruction error on the voxel grid inform where to initialize the next one (bottom). 
    }
    \label{fig:iter_example}
  \label{fig:test1}
\end{figure}

\begin{figure}
    \centering    
    \includegraphics[width=1.\linewidth]{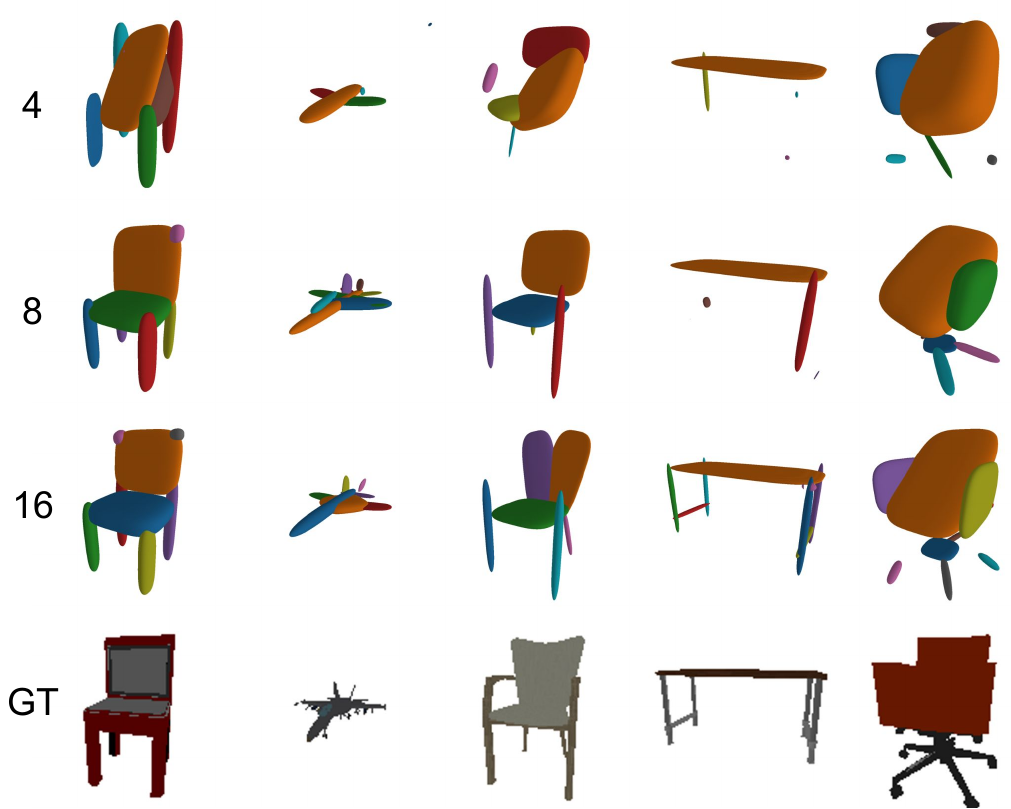}
    \captionof{figure}{\textbf{\ours\ with different number of views.} Objects are reconstructed with greater detail as we increase the views from 4 to 16, since they disambiguate more details.}
    \label{fig:views_ablation}
\end{figure}

\myparagraph{Number of views.}
As we add more views from different angles of the scene, our 3D recomposition becomes more accurate as reported in Table~\ref{tab:quan_views}. Even with as little as 4 views, our algorithm outperforms the average results of EMS (using point clouds) and \oursNI\ (at 16 views) from Tab.~\ref{tab:quan_shapenet} (\ie 0.576 vs 0.33 and 0.574 average IoU) and, with 8 views, the average results largely surpass also NBP (\ie +0.05). Nevertheless, at a low number of views, it becomes more probable that parts of the object stay occluded and this may lead to poor reconstructions. In Fig.~\ref{fig:views_ablation}, we illustrate some examples for which we observe much lower IoU, especially at 4 views. Details and thinner object parts such as \textit{chair} and \textit{table legs} more often contain errors or are not covered by superquadrics due to potential ambiguities on the 2D views.

 {
 \begin{table}
    \centering
\resizebox{1.0\columnwidth}{!}{
    \begin{tabular}{l|cc|cc|cc}
        Views & \multicolumn{2}{c|}{4} & \multicolumn{2}{c|}{8} & \multicolumn{2}{c}{16}\\
        
        Method & \oursNI & \ours & \oursNI & \ours & \oursNI & \ours\\
        \midrule
        airplane & 0.520 & 0.578  & 0.594 & 0.657      & 0.626 & \bf{0.696}\\
        bench    & 0.375 & 0.425  & 0.446 & 0.489      & 0.489 & \bf{0.499}\\
        cabinet  & 0.565 & 0.730  & 0.578 & \bf{0.746} & 0.599 & \bf{0.744}\\
        car      & 0.531 & 0.637  & 0.598 & 0.680      & 0.610 & \bf{0.703}\\ 
        chair    & 0.398 & 0.396  & 0.501 & 0.522      & 0.526 & \bf{0.549}\\
        display  & 0.517 & 0.533  & 0.607 & 0.599      & 0.629 & \bf{0.631}\\
        lamp     & 0.355 & 0.492  & 0.384 & 0.553      & 0.404 & \bf{0.531}\\
        speaker  & 0.544 & 0.709  & 0.554 & 0.746      & 0.576 & \bf{0.756}\\
        rifle    & 0.468 & 0.566  & 0.562 & 0.656      & 0.584 & \bf{0.692}\\
        sofa     & 0.529 & 0.577  & 0.600 & 0.666      & 0.601 & \bf{0.694}\\
        table    & 0.404 & 0.495  & 0.464 & 0.540      & 0.487 & \bf{0.576}\\
        phone    & 0.629 & 0.739  & 0.702 & 0.770      & 0.729 & \bf{0.791}\\
        vessel   & 0.569 & 0.611  & 0.645 & \bf{0.661} & 0.652 & \bf{0.663}\\
        \midrule                                                                                          
        mean     & 0.493  & 0.576 & 0.557 & 0.637  &  0.578  & \bf{0.656}\\
    \end{tabular}
   }
    \caption{\textbf{Analysis on the number of views.} We report IoU of the object volume on ShapeNet (higher is better). \ours\ consistently outperforms its non-iterative version \oursNI. 
    }
    \label{tab:quan_views}
\end{table}
 }

\myparagraph{Number of superquadrics.} \ours\ adds superquadrics one by one to the scene making the reconstruction gradually more detailed and complex. For explainability purposes, we would like to describe the object with few shapes that match the key parts of the object rather than fitting every detail with increasingly smaller superquadrics. In this regard, \ours\ is flexible to choose a subset of the maximum number of superquadrics to fit the target object, visualizing them after every iteration. Note that by focusing on local fitting with $\mathcal{L}_\lambda$ and 
initializing superquadrics at regions with high error, we obtain meaningful intermediate representations, 
even with few superquadrics. 

In Fig.~\ref{fig:iter_example}, we show how \ours\ builds objects iteratively from their constituent parts. For each object, the top row depicts the superquadrics reconstruction throughout the algorithm where each image adds one superquadric. The row below shows a rendering of the error map we obtain by backpropagating gradients onto the volumetric voxel grid,  guiding the placement of the next superquadric. We observe that coarse object parts are reconstructed first, with later superquadrics focusing on more fine-grained details, \eg \ours\ first represents the \textit{seat} and \textit{backrest} for the \textit{chair}, the \textit{body} and \textit{wings} for the \textit{airplane} and the \textit{table top}. 

\begin{figure}
    \centering
   \includegraphics[width=1.\linewidth]{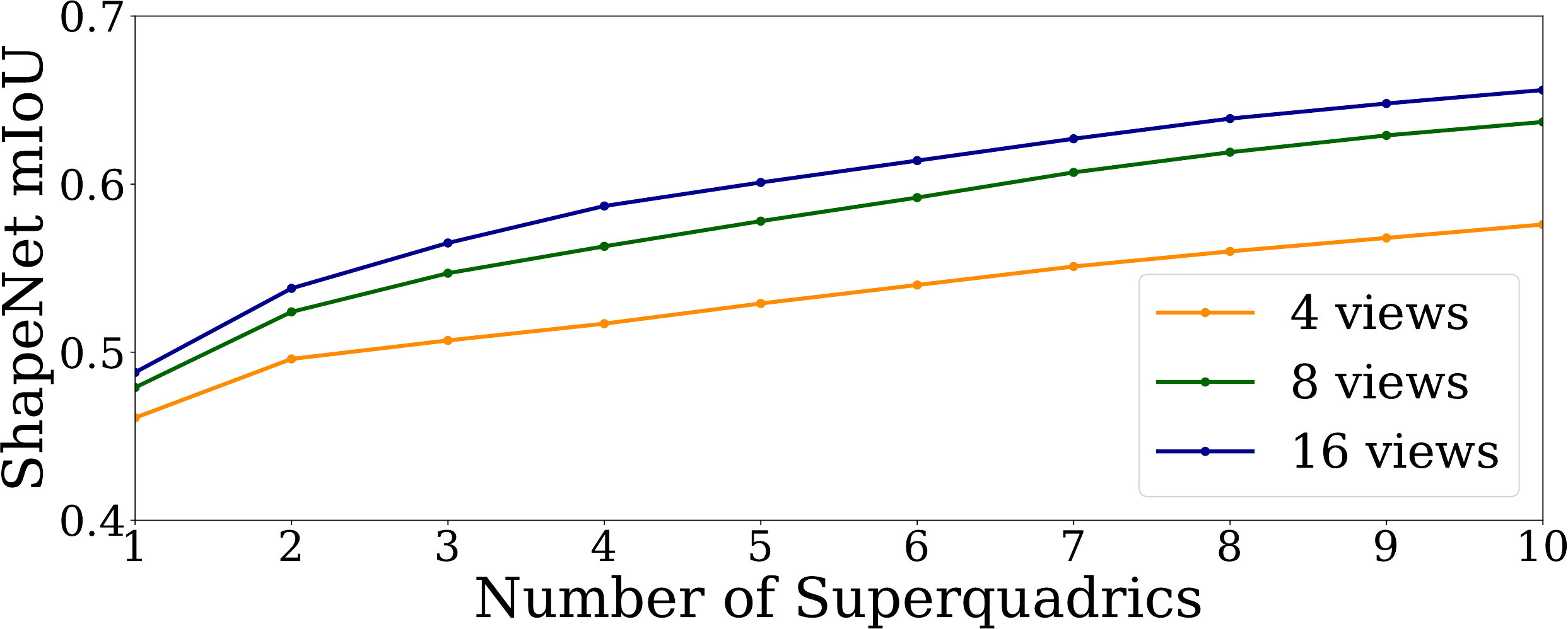}
    \caption{\textbf{Analysis on the number of superquadrics.} We report 
    the mIoU on ShapeNet for increasing number of superquadrics during \ours\ execution. Performance increase with the number of superquadrics. 
    }
    \label{fig:quan_nprimitive}
\end{figure}

From Fig.~\ref{fig:quan_nprimitive} we observe the gradual improvement in reconstruction fidelity (in IoU) as more superquadrics are added to the scene. With many superquadrics also comes diminishing returns at an additional cost of interpretability, which then contrasts with the goal of 
expose meaningful object parts, useful for downstream tasks. 
Therefore, we use a small number of superquadrics, \ie 10. 

\begin{figure}[t]
\includegraphics[width=1.0\linewidth]{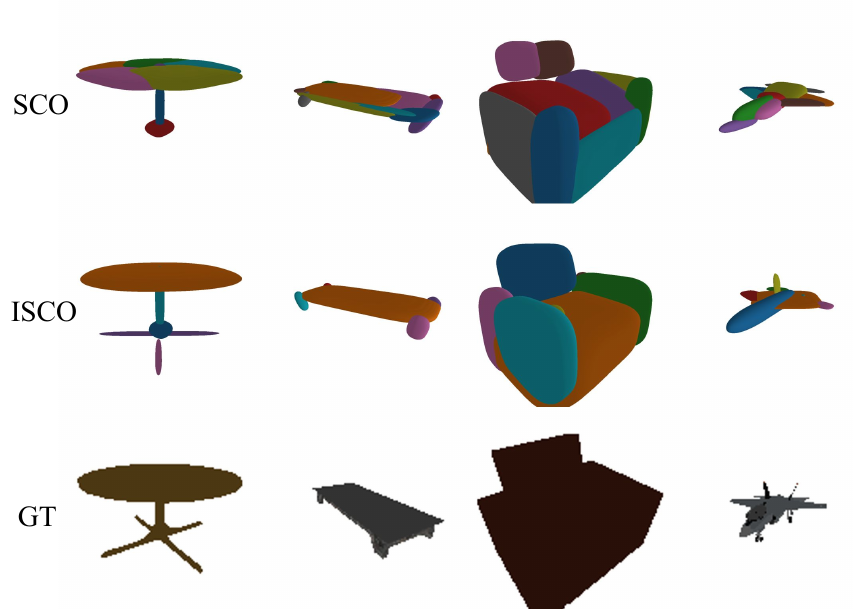}
   \caption{\textbf{The iterative procedure.} While \ours\ precisely covers multiple object parts, in \oursNI\ superquadrics compete on the same regions to less interpretable recompositions. }
    \label{fig:iterative_ablation}
\end{figure}

\myparagraph{Iterative fitting.} 
In Fig.~\ref{fig:iterative_ablation}, we show some examples of different objects, where key object parts are correctly covered by a single superquadric in \ours, and are made up of multiple ones in \oursNI. This harms the interpretability of individual shapes and their ability to semantically decompose an object. Moreover, details of the object are often not covered, i.e. legs of the \textit{chair}. In this case, the targeted initializations in \ours\ help reconstruct even fine details. 

These results are corroborated by the overall performance drop in Tab.~\ref{tab:quan_shapenet} and analyzed in Tab.~\ref{tab:quan_views} for 4 and 8 views. 
These qualitative and quantitative results demonstrate that, without the iterative procedure, superquadrics tend to compete to cover the whole object shape, without covering finer details. Our iterative procedure avoids this and allows each superquadric to focus on specific parts before introducing new elements in the scene. 

\subsection{Semantic Parsing on ShapeNet-parts}

\begin{table}[t]
\resizebox{1.\linewidth}{!}{
    \begin{tabular}{l|ccc|ccc}
& \multirow{2}{*}{MV} & \multirow{2}{*}{TF} & \multirow{2}{*}{C}&  plane & chair & table \\
&  &   &                                        & 4 parts & 4 parts  & 2 parts\\
\midrule
SQ \cite{paschalidou2019superquadrics}&  &   &  \cmark & 48.9        &  65.6         & 77.7 \\
BAE \cite{chen2019bae}&   &   & \cmark &    61.1     &     65.5      & 87.0 \\
RIM \cite{niu2022rim}&  &   & \cmark &   \textbf{67.8}      &      \textbf{81.5}     &  \textbf{91.2}\\
\midrule
\midrule
EMS \cite{liu2022robust}&  & \cmark  &  & 33.4        & 45.6          & 46.9\\
NBP \cite{wu2022primitive}&  & \cmark   &  &  33.8       & 53.4          & 44.3\\
\ours  &\cmark & \cmark  &  & \textbf{67.7}         & \textbf{76.6}       & \textbf{81.9}\\
\midrule
EMS \cite{liu2022robust}&  &  \cmark & \cmark &   26.0      &   38.1        & 32.3\\
NBP \cite{wu2022primitive}&  & \cmark  &  \cmark&     23.1    &     38.7      & 32.9\\
\ours & \cmark &  \cmark & \cmark &       \textbf{33.9}  &  \textbf{55.5}        & \textbf{70.7}\\
    \end{tabular}
    }
    \caption{\textbf{Per-part mIoU (in $\%$) on ShapeNet part~\cite{Yi16}.} Top: methods with point cloud input, using training set, and consistent part decomposition (C). Middle: training-set free (TF) and superquadric-based method with optimal semantic assignment. Bottom: the same method with fixed assignment. \ours\ uses multiple views (MV) as input.}
    \label{tab:quan_part}
\end{table}

We follow~\cite{niu2022rim} and conduct an analysis of recovering object parts using ShapeNet part~\cite{Yi16}, measuring the performance as mIoU between the ground-truth shape part and the superquadrics assigned to it. In Tab.~\ref{tab:quan_part}, we compare our model with EMS, NBP and the reference methods SQ~\cite{paschalidou2019superquadrics}, BAE~\cite{ChenYFCZ19} and RIM~\cite{niu2022rim}. While SQ is a shape abstraction method using superquadrics, BAE and RIM are semantic parsing ones. In the table, all three methods use point clouds (\eg as target in SQ, as input in BAE) 
and a training set of 3D shapes to achieve semantic parsing consistency. 

Since ours, EMS and NBP perform instance-based abstraction and cannot directly assign semantics to parts, we considered two strategies for assigning semantic to superquadrics: \textit{instance-wise}, where each superquadric is assigned to the closest object part, 
and \textit{consistent},  
where the order of the superquadrics determines the object part they are assigned to. In the latter case, the semantic identity of a given position is chosen as in~\cite{niu2022rim} for unsupervised objectives.

Despite using weaker 3D information, our method outperforms EMS and NBP by a margin for all categories and both with and without consistent semantic assignments (\eg with consistency, +8\% mIoU on \textit{plane}, +17\% mIoU on \textit{chair} and +38\% mIoU on \textit{table}). 
These results show the superiority of our model w.r.t. comparable superquadrics instance-based methods also in identifying semantic parts of the target shapes, even when consistency is enforced, showing the effectiveness of our iterative strategy reconstructing first coarse and then fine object regions. 

Even though \ours\ does not rely on learned per-class inductive biases as the competitors in the top part (\ie SQ, BAE, RIM) 
it still obtains relatively good results when consistency is enforced (\eg -7\% w.r.t. SQ on \textit{table}). The 
  results without consistent assignments even surpass the training set-based competitors (\eg +10\% on BAE and SQ on \textit{chair}) confirming the capability of \ours\ in detecting the most relevant object parts, even if not always maintaining the same ordering.  
In fact, since \ours\ is performed independently on each instance, there is no guarantee to obtain semantic consistency across instances of the same class, \eg for the \textit{plane} class the second and third superquadric could represent the left and right wing, respectively, but for another plane it might be the other way around. This causes the drop between the results with and without consistency.

\subsection{Shape Recomposition of Real-world images}

\begin{figure}
    \centering
    \includegraphics[width=1.0\linewidth]{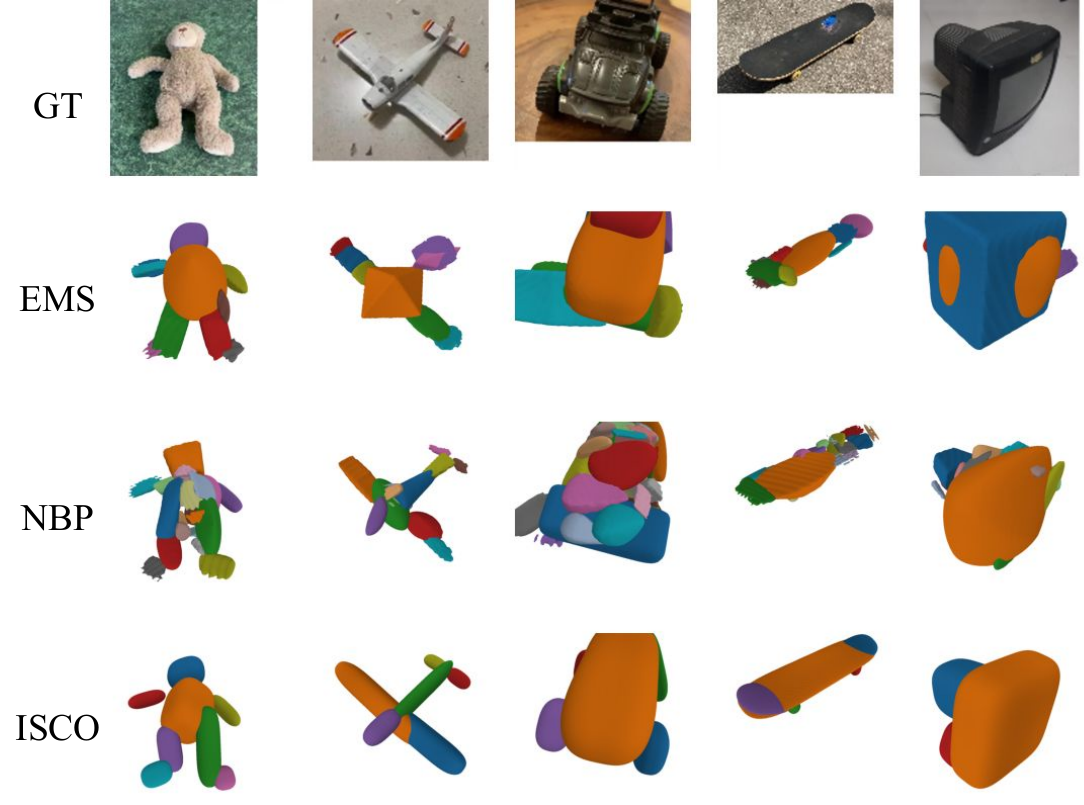}
    \caption{\textbf{Qualitative results on CO3D dataset.} We use 16 random views from each object, preprocessed to obtain their 
    instance masks. 
    Our method is the only one that accurately recomposes the target object and its parts.
    }
    \label{fig:co3d_qual}
\end{figure}

Using the Co3D, we can test \ours\ and the baselines on 
recomposition of objects in the 
real-world. 
In Fig.~\ref{fig:co3d_qual}, we present some qualitative examples of running \ours, EMS, and NBP on scenes from CO3D that contain a \textit{teddy bear}, a \textit{toy plane} and a \textit{toy truck}, a \textit{laptop}, a \textit{skateboard} and a \textit{tv}. We select these classes since they 
contain more complex shapes than others in the dataset (\eg 
\textit{apple}, \textit{ball}, \textit{book}).

CO3D poses several additional challenges not present in the synthetic ShapeNet. To start with, object masks are obtained via instance segmentation with PointRend~\cite{KirillovWHG20}. While the masks fit the object well on average, they contain noise and every set of images usually contains some artifacts, \eg when the background color blends easily with the object. These challenges extend to EMS and NBP, since reconstructed point clouds are more noisy from few views.  Nevertheless, \ours\ is generally robust to these type of noises, showing consistently better superquadrics recomposition than the competitors. Examples are the \textit{tv}, where EMS covers the main body with two overlapped superquadrics, or \textit{skateboard} where NBP uses several superquadrics to cover a small portion of the object. 
 
We highlight that, while neural networks can be used to learn shape priors from training data \cite{tulsiani2017learning,paschalidou2019superquadrics,paschalidou2021neural}, 
they to not generalize well to new shapes~\cite{zhang2018learning}. By performing instance-level training, \ours\ is not affected by changes on classes or input distributions: for instance, it can decompose the \textit{teddy bear} well into its parts, even if \eg ShapeNet does not contain a class with similar anatomy to learn from.

\section{Conclusion and Limitations}

As a conclusion, we present \oursFull\ (\ours), a self-supervised algorithm that recomposes 3D objects from multiple views using superquadrics. 
\ours\ uses a differentiable rendering pipeline to fit parameters of superquadrics to the target scene by comparing rendered with real silhouettes. Superquadrics are added once at the time to the scene, initializing them over the regions with the highest estimated reconstruction error. This guarantees that superquadrics explain a dedicated object region well, before additional ones are optimized to further improve coverage. Results on ShapeNet and ShapeNet-parts show that \ours\ is not only effective in reconstructing the target shape, but also in identifying its semantic parts, outperforming methods using point clouds as input. 
Lastly, we show the versatility of \ours\ by applying it on 
in-the-wild objects from natural images of CO3D.

While the results show the effectiveness of \ours, 
some acquisition conditions may impact the results. 
When views 
do not represent well the object, the reconstructions of \ours\ may fail, \eg the non-perfect wings of the \textit{toy-plane} in CO3D (Fig.~\ref{fig:co3d_qual}) due to lack of side views, or the ambiguous concavity of the rightmost \textit{chair} on Fig. \ref{fig:views_ablation}, that results in a unified seat/back. 
Extending \ours\ to include shape priors and/or 
reconstruct texture and lighting conditions can mitigate this limitation.
Another limitation 
is the computational cost. 
One update step on a 2080ti takes 13ms-22ms (for 4-16 views), obtaining a recomposition after 33s-55s. 
However, we did not optimize the runtime by \eg 
choosing the number of superquadrics by early stopping when the initialization step finds a low reconstruction error. 
Note also that NBP has a comparable running time (\eg 20-100s per instance but from point clouds) and that while EMS takes less than $1$s per instance it still requires extracting a point cloud for real-world applications. Advances on neural rendering can also be applied to \ours\ to further improve its runtime. 

\vspace{10pt}

\noindent\textbf{Acknowledgements.} This work was supported by ERC (853489 - DEXIM), BMBF (T\"ubingen AI Center, FKZ: 01IS18039A), DFG (2064/1 – project number 390727645), and by the MUR PNRR project FAIR - Future AI Research (PE00000013) funded by the NextGenerationEU.

{\small
\bibliographystyle{ieee_fullname}
\bibliography{egbib}
}

\newpage
\appendix
\title{Iterative Superquadric Recomposition of 3D Objects from Multiple Views\\-\\Supplementary Material}
\author{}
\maketitle

In this supplementary material, we provide a derivation on how we initialize superquadrics (Sec.~\ref{sec:init}), additional implementation details (Sec.~\ref{sec:implementation-details}), analyses on the effect of the hyperparameter $\lambda$ (Sec.~\ref{sec:lambda}) on our loss function (Eq.~6), and show how the camera viewpoints impact the performance of \ours\ 
(Sec.~\ref{sec:cameraviews}). We will then report quantitative results using Chamfer distance as metric (Sec.~\ref{sec:chamfer}), a comparison with trained abstraction methods (Sec.~\ref{sec:comparison}), and additional qualitative results (Sec.~\ref{sec:qual-shapenet}) on ShapeNet. 

\section{Superquadric initialization}
\label{sec:init}
To instantiate new superquadrics around object parts that have not been covered yet, we estimate the origin of rendering errors in the 3D space. We use a dense voxel grid $G \in \mathbb{R}^{N \times N \times N}$ with resolution $N$ around the object to which we propagate rendering errors.

First, we evaluate the superquadrics density at each point $\mathbf{g}$ in the voxel grid
\begin{equation}
    V_{\mathbf{g}} = \sigma(\mathbf{g}; \theta)
\end{equation}
where $V$ is the the complete grid of density values while $V_{\mathbf{g}}$ is the density value at $\mathbf{g}$. To simplify notation we omit $\theta$ from $V$ and any further definitions.
We then render this volume grid through ray marching and calculate the loss $\mathcal{L}^{\lambda=0}_{k-1}$. To do so, the density of each point along the camera ray $\mathbf{r}(t)$ is obtained by applying a sampling kernel locally
\begin{equation}
    K_{\mathbf{r}(t)} = \sum_{\mathbf{g} \in G} V_{\mathbf{g}} k(\mathbf{r}(t) - \mathbf{g}; \Phi)
\end{equation}
where $\Phi$ are the parameters of a generic sampling kernel $k(\cdot; \Phi)$ and we evaluate all ray points $\mathbf{r}(t)$ from the stratified sampling approach to render the reconstructed image.
In practice, we use trilinear interpolation for the kernel $k$ which results in
\begin{equation}
    K_{\mathbf{r}(t)} = \sum_{\mathbf{g} \in G} V_{\mathbf{g}} \prod_{i=1}^3 \max(0, 1 - \frac{|x_i^{(\mathbf{r}(t))} - x_i^{(\mathbf{g})}|}{l})
\end{equation}
where $[x_1^{(\mathbf{p})}, x_2^{(\mathbf{p})}, x_3^{(\mathbf{p})}]$ are the three-dimensional coordinates of point $\mathbf{p}$ and $l$ is the uniform distance between voxels in the grid.

Using the resampled density values from $K_{\mathbf{r}(t)}$ to replace the true densities of the superquadrics, we can now render the reconstructed image with
\begin{equation}
    D(\textbf{r}) = \int_{t_n}^{t_f} T(t) K_{\mathbf{r}(t)}dt
\end{equation}
and calculate the loss $\mathcal{L}^{\lambda=0}_{k-1}$ the same way as described in the paper for direct rendering of the superquadrics.

Finally, to propagate the rendering errors, we calculate the gradient of the loss with respect to the densities $V_{\mathbf{g}}$ of the grid points
\begin{equation}
    \frac{\partial {\mathcal{L}_{k-1}^0}}{\partial V_{\mathbf{g}}} = \sum_{\textbf{r}\in \mathcal{R}} \sum_{\textbf{r}(t) \in \textbf{r}} \frac{\partial \mathcal{L}_{k-1}^0}{\partial{K_{\textbf{r}(t)}}} \prod_{i=1}^3 \max(0, 1 - \frac{|x_i^{(\mathbf{r}(t))} - x_i^{(\mathbf{g})}|}{l})
\end{equation}
where $\frac{\partial \mathcal{L}_{k-1}^0}{\partial{K_{\textbf{r}(t)}}}$ is calculated as is common in ray marching and the remaining part of the term is a result of the trilinear sampling that distributes the gradient to each point $\mathbf{g}$ (using $V_{\mathbf{g}}$ as a representative value for its density contribution) in the voxel grid $G$.

Intuitively, each point $\mathbf{g}$ is a candidate location for initializing a new superquadric. Hence, we estimate how accurate the density at each location $\mathbf{g}$ is. We use $\lambda=0$ in $\mathcal{L}^{\lambda=0}_{k-1}$ to only consider errors where part of the object is not covered by a superquadric yet. By propagating the error to each $\mathbf{g}$, the point with the highest error corresponds to the location where we have the best potential to improve our reconstruction loss if we initialize a new superquadric there.

We choose the voxel grid resolution $N=64$ such that there are $64^3 \approx 262\text{k}$ candidate locations for new superquadrics in each \ours\ iteration. The voxel grid is placed in the center of the scene (where we expect the object to be) and the distance $l$ between voxels is chosen such that the object is completely enclosed by the voxel grid. Since the exact position and size of the object are unknown, we choose $l$ based on the distance between the cameras and the center of the scene.

\section{Implementation details}
\label{sec:implementation-details}
\textbf{Learning rate.}  We found that fitting the superquadrics to the ground-truth 2D views, can be sensitive to the learning rate during optimization. Specifically, when the learning rate is too small, significantly more update steps are required to fit the superquadric parameters well, before introducing the next superquadric. We found that a learning rate of $0.01$ helps in optimizing the first superquadrics reliably within our chosen 250 update steps per iteration. However, as smaller regions of the shape are being reconstructed later on, more fine-grained optimization steps are necessary, and thus we gradually reduce the learning rate to $0.001$. Hence, for all our experiments, we use the Adam~\cite{kingma2015adam} optimizer and a cosine learning rate schedule that starts with a learning rate of $0.01$ and is annealed to $0.001$ by the end of the optimization. All the hyperparameters have been selected using a small subset of object instances in ShapeNet training set and kept constant across all the experiments.

\textbf{Ray sampling.} Due to computational reasons, it is not common to render the whole image from all camera angles during training, but instead subsample rays, i.e. pixels, for every viewpoint \cite{mildenhall2020nerf}. Since our loss is defined on object silhouettes, a large amount of pixels will fall outside both the target object and the superquadrics rendering, not contributing to the loss. Hence, we employ an importance sampling strategy, by sampling rays that contributed to the loss in previous update steps with higher probability. This improves sampling efficiency and allows us to use 500 rays per view point rather than all and 250 update steps per superquadric.

\section{Choice of $\lambda$}
\label{sec:lambda}
The hyperparameter $\lambda$ in our loss $\mathcal{L}(\lambda)$ regulates local vs. global loss terms. For instance, when fitting a single superquadric to the object, $ 0 < \lambda < 0.5$ encourages the superquadric to fully enclose the object shape and promoting global fitting, while $\lambda > 0.5$ produces tighter boundaries, promoting local fitting. As Table \ref{tab:quan_lambda} shows, performance rapidly decrease if $\lambda$ is either too low (\eg $\lambda=0.4$) or too high (\eg $\lambda=0.8$) since either the model cover larger regions (even beyond the target object) or it focuses on too fine-grained details, even ignoring other object parts.   

Quantitatively, we observed that a larger $\lambda$ (\ie higher focus on locality) lead to a better locality and accuracy in decomposing objects into meaningful parts,
with $\lambda=0.6$ leading to the best overall performance by $\sim 1$\% IoU on ShapeNet, cf. Table~\ref{tab:quan_lambda}. Based on these observations, we choose $\lambda=0.6$ throughout the experiments in this work.

\begin{table}
    \centering
    \begin{tabular}{c|ccccc}
        & \multicolumn{5}{c}{$\lambda$}\\
        \# views & 0.4 & 0.5 & 0.6 & 0.7 & 0.8\\
        \midrule
        4   & 0.547 & 0.560 & \bf{0.569} & 0.551 & 0.541\\
        8   & 0.606 & 0.629 & \bf{0.630} & 0.615 & 0.593\\
        16   & 0.629 & 0.643 & \bf{0.650} & 0.631 & 0.613\\
    \end{tabular}
    \vspace{3pt}
    \caption{\textbf{Volumetric IoU on ShapeNet} of \ours\ for different values of $\lambda$ and different number of views. We measure the results as mean Intersection over Union (IoU) of the object volume on ShapeNet (higher is better).
    }
    \label{tab:quan_lambda}
\end{table}

\section{Dependence on camera viewpoints}
\label{sec:cameraviews}
In the main paper, camera views are sampled randomly around the object. If, in practice, this is not feasible, information about the object from certain angles might be missing. We investigate how such a constraint could limit the reconstruction accuracy on the ShapeNet dataset where we can control the choice of camera views best. If we constrain the 16 random views to be inside a spherical cap with polar angle 90\degree (top hemisphere), 45\degree and 22.5\degree (where 0\degree is the top-view), the reconstruction mIoU decreases from 0.656 to 0.634, 0.592 and 0.528, respectively. Despite the limited drop, these results show that \ours\ depends on the choice of camera views.

 {
 \begin{table}
    \centering
    \resizebox{0.8\columnwidth}{!}{
    \begin{tabular}{l|cc|c}
        Input &\multicolumn{2}{c}{Point Cloud}&\multicolumn{1}{c}{2D Views}\\
        Method & EMS~\cite{liu2022robust} & NBP~\cite{wu2022primitive} & \ours\\
        \midrule
        airplane & 0.1201 & 0.0590 & \bf{0.0508}\\
        bench    & 0.2105 & 0.0725 & \bf{0.0611}\\
        cabinet  & 0.1327 & \bf{0.0990} & 0.1097\\
        car      & 0.0743 & 0.0730 & \bf{0.0530}\\ 
        chair    & 0.2225 & 0.1303 & \bf{0.1259}\\
        display  & 0.1353 & 0.1009 & \bf{0.0978}\\
        lamp     & 0.2012 & 0.1338 & \bf{0.1259}\\
        speaker  & 0.2098 & 0.1775 & \bf{0.1428}\\
        rifle    & 0.1050 & 0.0577 & \bf{0.0191}\\
        sofa     & 0.1684 & 0.0952 & \bf{0.0807}\\
        table    & 0.2467 & 0.1649 & \bf{0.1418}\\
        phone    & 0.0634 & 0.0406 & \bf{0.0350}\\
        vessel   & 0.0987 & 0.0468 & \bf{0.0393}\\
        \midrule                                                                                           
        mean     & 0.1530 & 0.0962 & \bf{0.0833}\\
    \end{tabular}
    }
    \vspace{3pt}
    \caption{\textbf{Chamfer-L1 Distance on ShapeNet.} We report Chamfer-L1 distance on ShapeNet (lower is better). EMS and NBP use point cloud as input, ours uses 16 random views.
    }
    \label{tab:chamfer_shapenet}
\end{table}

 \section{Chamfer-L1 evaluation on ShapeNet}
\label{sec:chamfer}
 In addition to the IoU reconstruction results in the main paper, here we report the Chamfer-L1 distance between the reconstructed superquadrics shapes of EMS~\cite{liu2022robust}, NBP~\cite{wu2022primitive} and \ours\ to the ground-truth shapes in Table~\ref{tab:chamfer_shapenet}. We sample 100k points on the surface of both the ground truth and predicted shape to calculate the Chamfer-L1 distance.  Note that in Table~\ref{tab:chamfer_shapenet} 
 \ours\ uses multi-view 
 inputs, while EMS and NBP 
 points clouds are extracted from the target 3D object. 
 
 In all but one category, the superquadric reconstructions of \ours\ are closer to the original shapes than both EMS and NBP. This result follows the same trend as the IoU (Table 1 of the main paper), providing further evidence that \ours\ can better recompose objects with superquadrics than its competitors, despite relying on cheaper multiple-view inputs rather than ground-truth 3D representations.

\section{Comparison with unsupervised abstraction methods using a training set}
\label{sec:comparison}

Shape abstraction has been studied by performing single-view reconstruction with a neural network. These methods typically use a single view as input and dense 3D point clouds at targets. By performing training on a large dataset, these networks learn shape priors such that they can reconstruct an object with simple shapes from an image. Related works include SQ~\cite{paschalidou2019superquadrics} and HSQ~\cite{paschalidou2020learning} which also use superquadrics as primitive shapes. However, since \ours\ takes a different instance-based perspective without requiring a training set, it is difficult to make a comparison on the same ground. For instance, SQ and HSQ, both single-view but with a training set, achieve 0.277 and 0.580 mIoU on ShapeNet, respectively, while ours 0.656 mIoU, not trained but with multi-views. We tried to perform a fair assessment, training HSQ with 16 views as input on all ShapeNet classes and also evaluating it with the same 16 views as \ours. In this case, the mIoU of HSQ improves slightly by 0.029, with ISCO still outperforming it. Note that, unlike our instance-based method, SQ and HSQ suffer from distribution shifts as they cannot generalize beyond the training set.

\section{Additional qualitative examples}
\label{sec:qual-shapenet}

In Figure~\ref{fig:classes}, we show additional qualitative results of the classes from ShapeNet where we randomly sample instances from the test set. We use the same hyper-parameters described in the main paper, using 10 superquadrics to represent the instances. From the results, we can see some failure cases of our model. For instance, thin parts may not be well decomposed, especially if they are hard to model with 2D views only, such as the body/tube of \textit{lamps} (4th row) or legs of \textit{benches} (2nd row).

At the same time, the figure also shows the ability of \ours\ to capture the structure of the underlying object. For instance, objects with distinct parts consistently have these decomposed by \ours\ such as the wings and engines of \textit{airplanes} (top row), body of \textit{vessels} (last row), and legs, seats, and back rests of
\textit{chairs}. For simpler shapes (e.g. \textit{phones}, 2nd from bottom), superquadrics are used in later stages to fill inaccuracies in corners of the object and smaller details. Note that if higher abstraction is desired, one could reduce the number of superquadrics, or prune small superquadrics post-hoc.

\begin{figure*}
    \centering
    \includegraphics[width=0.85\linewidth]{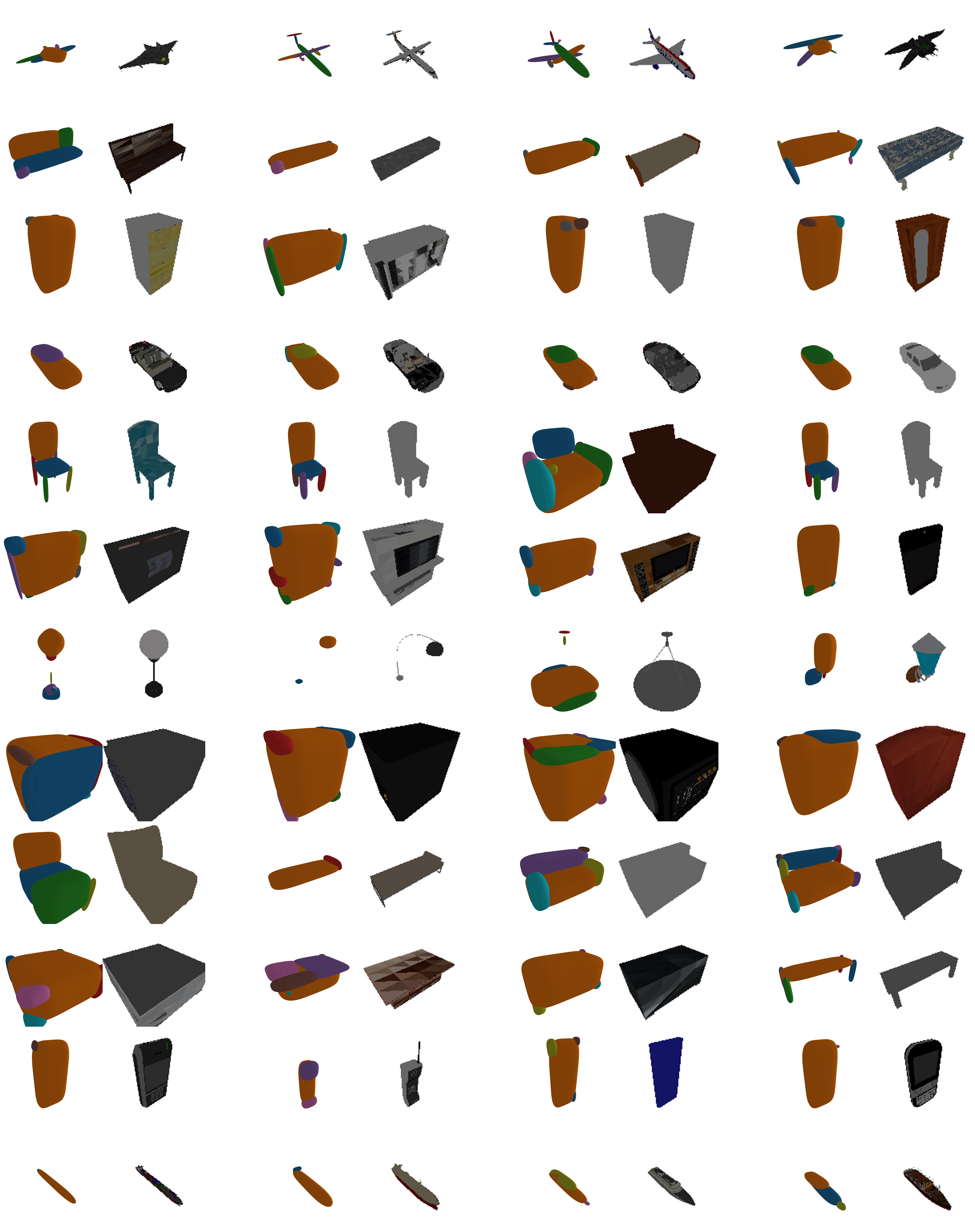}
    \caption{\textbf{Additional qualitative results for ShapeNet classes.} We show additional qualitative examples for \textit{random} instances of ShapeNet classes. Each row is a different class, from top to bottom:  \textit{airplane}, \textit{bench}, \textit{cabinet} , \textit{car}, \textit{chair}, \textit{display}, \textit{lamp}, \textit{speaker}, 
    \textit{sofa}, \textit{table}, \textit{phone}, \textit{vessel}. On each column, the left parts shows the result of our model while the right part the ground-truth shape.
    }
    \label{fig:classes}
\end{figure*}


\end{document}